\documentclass{llncs} 
\usepackage{cancel}

\newcommand{\resp}{\textit{resp.}~}
\newcommand{\projm}{proj_{max}}
\newcommand{\comp}{\; comp \; }
\newcommand{\spec}{\; spec \; }
\newcommand\lto{\mathbin{\backslash}}
\newcommand\lfrom{\mathbin{/}}
\newcommand\seq\vdash

\input{qobitree}
\usepackage{prooftree}
\usepackage[applemac]{inputenc}
\usepackage[pdftex]{graphicx}
\usepackage{amssymb}

\newtheorem{defi}{Definition}

\title{Minimalist Grammars and Minimalist Categorial Grammars, definitions toward inclusion of generated languages}
\titlerunning{MG and MCG, definitions}
\author{Maxime Amblard \inst{1}} 
\authorrunning{M. Amblard} 
\institute{Nancy Universit\'e - INRIA Nancy-Grand Est\\ \email{amblard@loria.fr}}

\begin{document}

\maketitle 
\begin{abstract} 
Stabler proposes an implementation of the Chomskyan Minimalist
Program, \cite{Chom95} with Minimalist Grammars - MG, \cite{Sta97}.
This framework inherits a long linguistic tradition.  But the semantic
calculus is more easily added if one uses the Curry-Howard
isomorphism.  Minimalist Categorial Grammars - MCG, based on an
extension of the Lambek calculus, the mixed logic, were introduced to
provide a theoretically-motivated syntax-semantics interface,
\cite{AM07th}.  In this article, we give full definitions of MG with
algebraic tree descriptions and of MCG, and take the first steps
towards giving a proof of inclusion of their generated languages.
\end{abstract}

The Minimalist Program - MP, introduced by Chomsky, \cite{Chom95},
unified more than fifty years of linguistic research in a theoretical
way.  MP postulates that a \emph{logical form} and a \emph{sound}
could be derived from \emph{syntactic relations}. Stabler,
\cite{Sta97}, proposes a framework for this program in a computational
perspective with Minimalist Grammars - MG. These grammars inherit a
long tradition of generative linguistics.  The most interesting
contribution of these grammars is certainly that the derivation system
is defined with only two rules: \emph{merge} and \emph{move}.  The
word \textit{Minimalist} is introduced in this perspective of
simplicity of the definitions of the framework. If the \textit{merge}
rule seems to be classic for this kind of treatment, the second rule,
\textit{move}, accounts for the main concepts of this theory and makes
it possible to modify relations between elements in the derived
structure.

Even if the phonological calculus is already defined, the logical one
is more complex to express.  Recently, solutions were explored that
exploited Curry's distinction between tectogrammatical and
phenogrammatical levels; for example, Lambda Grammars, \cite{RMusk03},
Abstract Categorial Grammars, \cite{PdG01b}, and Convergent Grammars \cite{Polletal09}.
First steps for a convergence between the Generative Theory and
Categorial Grammars are due to S. Epstein, \cite{BE96}. A full volume
of \textit{Language and Computation} proposes several articles in this
perspective, \cite{RLC}, in particular \cite{Lec04}, and Cornell's
works on links between Lambek calculus and Transformational Grammars,
\cite{tC04}.  Formulations of Minimalist Grammars in a Type-Theoretic
way have also been proposed in \cite{LecR99}, \cite{LR01acl},
\cite{AL05}.  These frameworks were evolved in \cite{ALR03},
\cite{AM07th}, \cite{ALR04} for the syntax-semantics interface.

Defining a syntax-semantics interface is complex. In his works,
Stabler proposes to include this treatment directly in MG. But
interactions between syntax and semantic properties occur at different
levels of representation.  One solution is to suppose that these two
levels should be synchronized. Then, the Curry-Howard isomorphism
could be invoked to build a logical representation of utterances.  The
Minimalist Categorial Grammars have been defined from this perspective:
capture the same properties as MG and propose a synchronized semantic
calculus. We will propose definitions of these grammars in this
article.  But do MG and MCG 
genrate the same language?
In this article we take the first steps towrds showing that they do.

The first section proposes new definitions of Minimalist Grammars
based on an algebraic description of trees which allows to check
properties of this framework, \cite{AM07th}.  In the second section,
we will focus on full definitions of Minimalist Categorial Grammars
(especially the phonological calculus). We will give a short
motivation for the syntax-semantics interface, but the complete
presentation is delayed to a specific article with a complete example.
These two parts should be viewed as a first step of the proof of
mutual inclusion of languages between MG and MCG. This property is
important because it enables us to reduce MG's to MCG, and we have a
well-defined syntax-semantics interface for MCG.

\section{ Minimalist Grammars}
Minimalist Grammars were introduced by Stabler \cite{Sta97} to
encode the Minimalist Program of Chomsky, \cite{Chom95}.  They capture
linguistic relations between constituents and build trees close to
classical Generative Analyses.

These grammars are fully lexicalized, that is to say they are
specified by their lexicon.  They are quite different from the
traditional definition of lexicalized because they allow the use of
specific items which do not carry any phonological form.  The use of
theses items implies that MG represent more than syntactic relations
and must be seen as a meta-calculus lead by the syntax.

These grammars build trees with two rules: \emph{merge} and
\emph{move} which are trigged by features.  This section presents all
the definitions of MG in a formal way, using algebraic descriptions of
trees.

\subsection{Minimalist Tree Structures\label{StructArbMin}}

To provide formal descriptions of Minimalist Grammars, we differ from
traditional definitions by using an algebraic description of trees: a
sub-tree is defined by its context, as in \cite{journals/jfp/Huet97}
and \cite{LC}. For example, the figure on the left of the figure
\ref{dominance} shows two subtrees in a tree ($t_1$ and $t_2$) and
their context ($C_1$ and $C_2$). Before we explain the relations in
minimalist trees, we give the formal material used to define a tree by
its context.

\subsubsection{Graded alphabets and trees:}

Trees are defined from a \emph{graded set}.  A graded set is made up
of a support set, noted $\Sigma$, the alphabet of the tree, and a
\emph{rank} function, noted $\sigma$, which defines node arity (the
\emph{graded} terminology results from the rank function).  In the
following, we will use $\Sigma$ to denote a graded $(\Sigma, \sigma)$.
	
The set of trees built on $\Sigma$, written $T_ {\Sigma}$, is the
smallest set of strings $ (\Sigma \cup \{(;);, \})^{\ast}$. A leaf of
a tree is a node of arity $0$, denoted by $\alpha$ instead of $ \alpha
()$.  For a tree $t$, if $t = \sigma (t_1, \cdots, t_k)$, the root
node of $t$ is written $\sigma$ .

Moreover, a set of variables $X = \{x_1, x_2, \cdots \}$ is added for
these trees.  $X_{k}$ is the set of $k$ variables.  These variables
mark positions in trees.  By using variables, we define a substitution
rule: given a tree $t \in T_{\Sigma(X_k)}$ (\textit{i.e.} a tree which
contains instances of $k$ variables $x_1, \cdots , x_k$) and $t_1,
\cdots, t_k$, $k$ trees in $T_{\Sigma}$, the tree obtained by
simultaneous substitution of each instance of $x_1$ by $t_1$, \dots,
$x_k$ by $t_k$ is denoted by $t[t_1, \cdots, t_k]$.  The set of all
subtrees of $t$ is noted $\mathcal{S}_t$.

Thus, for a given tree $t$ and a given node $n$ of $t$, the subtree
for which $n$ is the root is denoted by $t$ with this subtree replaced
by a variable.

Minimalist trees are produced by Minimalist Grammars and they are built on the graded alphabet $\{ <,>, \Sigma \}$, whose ranks of $ <$ and $>$ are $2$ and $0$ for strings of $\Sigma$. Minimalist Trees are binary ones whose nodes are labelled with $<$ or $>$, and whose leaves contain strings of $\Sigma$.

\subsubsection{Relations between sub-trees}
We formalise relations for different positions of elements in $\mathcal{S}_t$. Intuitively, these define the concept of \emph{be above}, \emph{be on the right} or \emph{on the left}. 
A specific relation on minimalist trees is also defined: \emph{projection} that introduces the concept of \emph{be the main element} in a tree.

In the following, we assume a given graded alphabet $\Sigma$. 
Proofs of principal properties and closure properties are all detailed in \cite{AM07th}.
The first relation is the dominance which informally is the concept of \textit{be above}.

\begin{defi}
Let $t \in T_\Sigma$, and $C_1, C_2 \in \mathcal{S}_t$, $C_1$ \textbf{dominates} $C_2$ (written $C_1 \lhd^{\ast} C_2$) if there exists $C' \in \mathcal{S}_t$ such that $C_1[C'] = C_2$.
\end{defi}

\begin{figure}[htbp]
\begin{center}
\begin{tabular}{ccc}
\includegraphics[height = 5 cm]{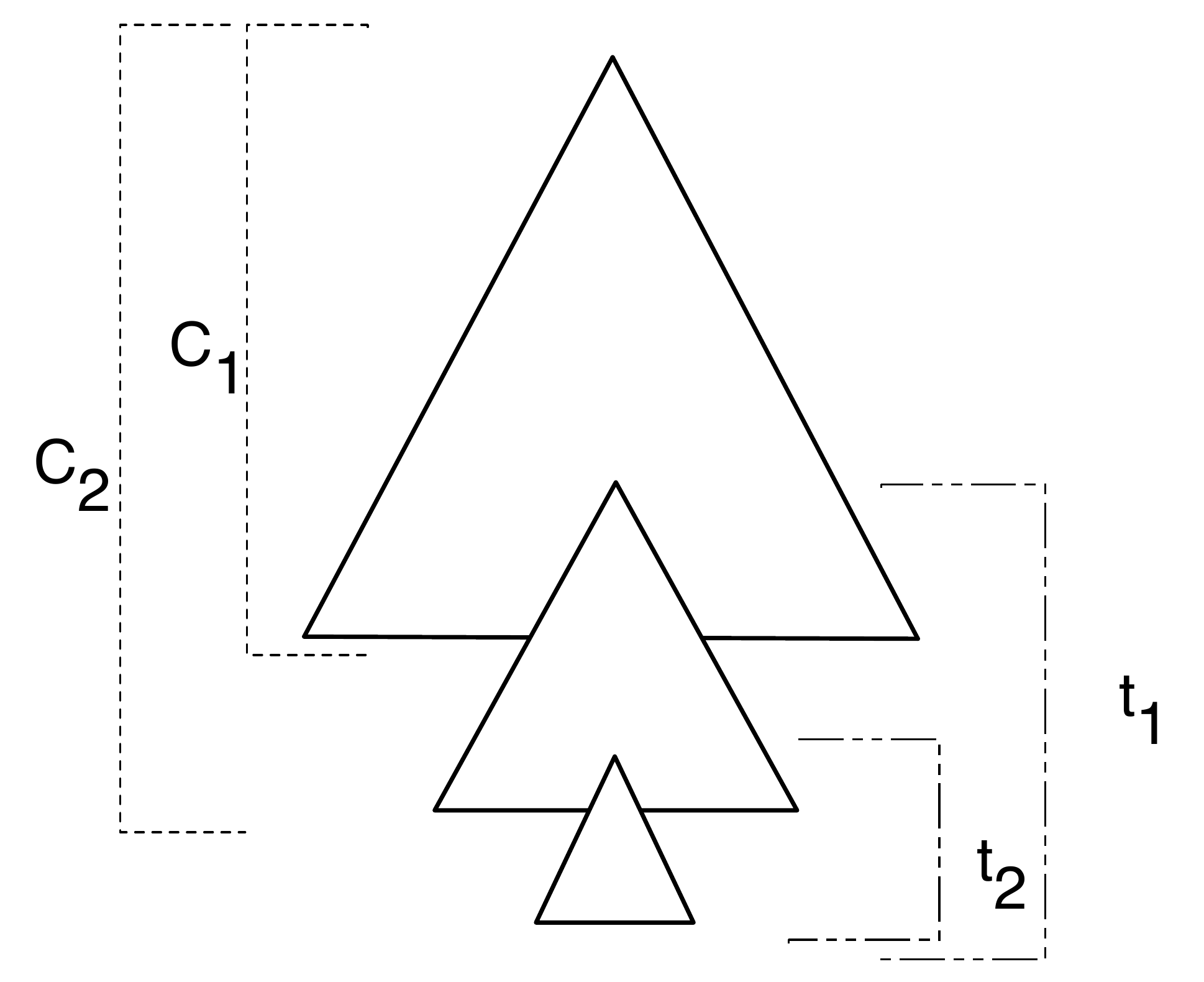}&$\quad$&
\includegraphics[height = 5 cm]{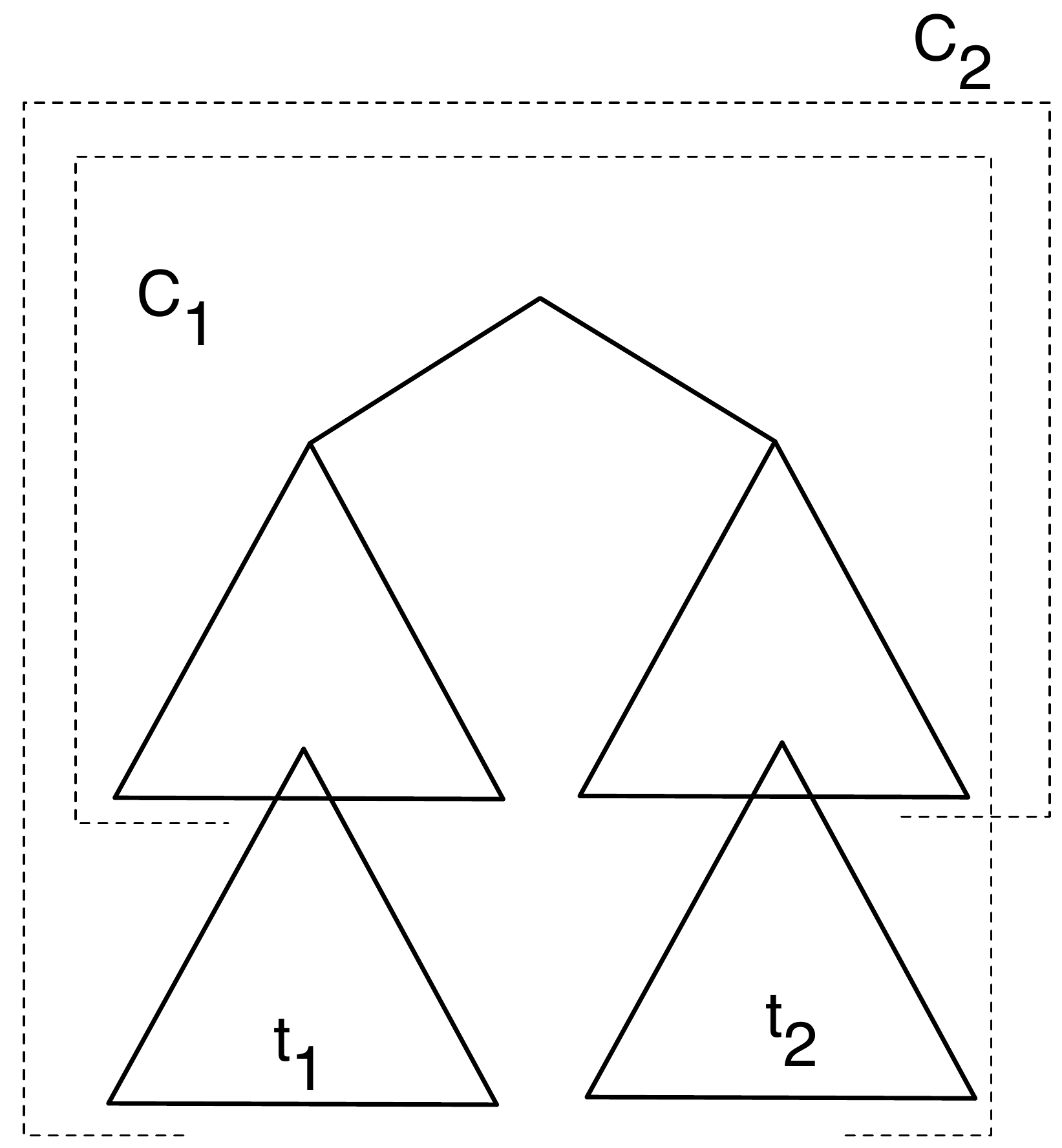}\\

\\
\begin{minipage}{6cm}
\begin{itemize}
\item $C_1$ is the context of the sub-tree $t_1$
\item $C_2$ is the context of the sub-tree $t_2$
\item $C_1 \lhd^{\ast} C_2$ means that the root node of $t_1$ is higher than the root node of $t_2$ in the full tree
\end{itemize}
\end{minipage}
&
&
\begin{minipage}{6cm}
\begin{itemize}
\item $C_1$ is the context of the sub-tree $t_1$
\item $C_2$ is the context of the sub-tree $t_2$
\item $C_1 <^{\ast} C_2$ means that the root node of $t_1$ is on the left side of the root node of $t_2$ in the tree
\end{itemize}
\end{minipage}

\end{tabular}
\end{center}
\caption{Dominance and precedence relations in trees\label{dominance}.}
\end{figure}

Figure \ref{dominance} shows an example of dominance in a tree.
One interesting property of this algebraic description of trees is that properties in sub-trees pass to tree.
For example, in a given tree $t$, if there exists $C_1$ and $C_2$ such that $C_1\lhd^{\ast} C_2$, using a 1-context $C$, we could build a new tree $t' = C[t]$ (substitution in the position marked by the variable x$x_1$ of $t$). Then, $C[C_1]$ and $C[C_2]$ exist (they are part of $t'$) such that $C[C_1]\lhd C[C_2]$.

\begin{defi}
Let $t \in T_\Sigma$, $C_1, C_2 \in \mathcal{S}_t$, $C_1$ \textbf{immediately precedes} $C_2$ (written $C_1 \prec C_2$) if there exists $C \in \mathcal{S}_t$ such that:
\begin{enumerate}
\item $C_1=C[\sigma(t_1,\ldots,t_j,x_1,t_{j+2},\ldots, t_k)]$ and
\item $C_2=C[\sigma(t_1,\ldots,t_j,t_{j+1}, x_1,\ldots, t_k)]$.
\end{enumerate}

\textbf{Precedence}, written $\prec^{\sim}$, is the smallest relation defined by the following rules (transitivity rule, closure rule and relation between dominance and precedence relation):

\begin{center}
\prooftree
	C_1 \prec^{\sim} C_2
	\quad C_2 \prec^{\sim} C_3
	\justifies
	C_1 \prec^{\sim} C_3
	\using[trans]
\endprooftree
\qquad
\prooftree
	C_1 \prec C_2
	\justifies
	C_1 \prec^{\sim} C_2
	\using[\ast]	
\endprooftree
\qquad
\prooftree
	C_1 \lhd^* C_2
	\justifies
	C_2 \prec^{\sim} C_1
	\using[dom]
\endprooftree
\end{center}

%closure of $\prec$, the transitivity rule and the following rule: if $C_1 \lhd^\ast C_2$ then $C_2 \prec^\sim C_1$.
\end{defi}

Precedence encodes the relation \textit{be on the left} (and then \textit{be on the right}) or \textit{be above} another element (using the dominance).
These two relations stay true for substitution (as mentioned above).

The next relation does not define a tree relation. It realises a linguistic property by leading the concept of \textit{be the main} element in a structure (or a substructure).

\begin{defi}
Let $t \in T_{\Sigma_{MG}}(A)$, and $C_1, C_2 \in \mathcal{S}_t$, $C_1$ \textbf{immediately projects} on $C_2$ (written $C_1 < C_2$) if there exists $C \in \mathcal{S}_t$ such that one of the two following properties holds:
\begin{enumerate}
\item $C_1 = C[{<}(x_1, t_2)]$ and $C_2 = C[{<}(t_1, x_1)]$,
\item $C_1 = C[{>}(t_2, x_1)]$ and $C_2 = C[{>}(x_1, t_1)]$,
\end{enumerate}

in this case $C \lhd C_1$ and $C \lhd C_2$. 
If $C_1 {<} C_2$ or $C_2 {<} C_1$, then there exists $C$ such that $C \lhd C_1$ and $C \lhd C_2$.

${<}^{\sim}$ is the smallest relation defined by the following system of rules:

\begin{center}
\prooftree
	C \in \mathcal{S}_t
	\justifies
	C {<}^{\sim} C
	\using[0]
\endprooftree
\qquad
\prooftree
	C_1 {<}^{\sim} C_2
	\quad C_2 {<}^{\sim} C_3
	\justifies
	C_1 {<}^{\sim} C_3
	\using[trans]
\endprooftree
\qquad
\prooftree
	C_1 {<} C_2
	\justifies
	C_1 {<}^{\sim} C_2
		\using[\sim]
\endprooftree

\bigskip

\prooftree
	C_1 \lhd^{\ast} C_2
	\quad C_3 \lhd^{\ast} C_4
	\quad C_2 {<} C_3
	\justifies
	C_1 {<}^{\sim} C_4
	\using[A]
\endprooftree
\qquad
\prooftree
	C_1 \lhd C_2
	\quad C_2 {<} C_3
	\justifies
	C_2 {<}^{\sim} C_1
	\using[B]	
\endprooftree
\end{center}

\end{defi}

Note that the projection relation is transitive. All the properties of these three relations are proven in  \cite{AM07th}. The figure \ref{exmintree} presents three minimalist trees where in $t$ the \textit{main} element is the verb \textit{walks} (which is accessible by following the projection relation).

These three relations could seem quite complicated for a reader who is not familiar with these notations or the zipper theory. But their expressiveness allows to prove the structural properties assumed for MG and moreover to give the proof of languages inclusion with MCG. Finally, in this section, we have defined the concept of parent and child relations in trees plus the projection relation which defines constituents in linguistic descriptions.

\subsection{Linguistic Structures in Trees}
From the linguistic perspective, trees represent relationships between grammatical elements of an utterance. Linguistic concepts are associated with minimalist tree structures.	
These relationships have been proposed for the analysis of structural analogies between verbal and nominal groups. 
Thus, groups of words in a coherent statement (phrases), whatever their nature, have a similar structure. This is supposed to be the same for all languages, regardless of the order of sub-terms. This assumption is one of the basic ideas of the $ X $-bar theory  introduced in the seventies, \cite{Chomsky73} and in the MP, \cite{Chom95}.\\
	
\subsubsection{The head}
is the element around which a group is composed.
An easy way to find the head of a minimalist tree is to follow the projection relation of the nodes.

\begin{defi}
Let $t \in T_{MG}$, if for all $C' \in \mathcal{S}_t, \; C {<}^{\sim} C'$ then $C$ is called the \textbf{head} of $t$.
For a given tree $t \in T_{MG}$, we write $H_t[x] \in S_t$ a sub-tree of $t$ of which $x$ is the head, and $head(t)$ is a leaf which is the head of $t$.
Then $t = H_t[head(t)]$.
\end{defi}

For a minimalist tree, there always exists a unique minimal element for the projection relation and it is a leaf (which is the head of the tree) \cite{AM07th}.\\ 

\begin{figure}
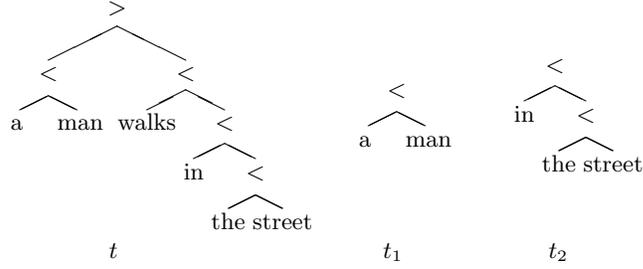

\vspace{-5ex}
\begin{tabular}{ccc}
\leaf{$\;$a$\;\;$}
\leaf{$\;\;$man$\;$}
\branch{2}{$<$}
\leaf{walks}
\leaf{in}
\leaf{the}
\leaf{street}
\branch{2}{$<$}
\branch{2}{$<$}
\branch{2}{$<$}
\branch{2}{$>$}
\tree
&
\leaf{$\;$a$\;\;$}
\leaf{$\;\;$man$\;$}
\branch{2}{$<$}
\tree
&
\leaf{in}
\leaf{the}
\leaf{street}
\branch{2}{$<$}
\branch{2}{$<$}
\tree\\
$t$ & $t_1$ & $t_2$\\
\end{tabular}

\caption{a minimalist tree $t$  and two of its sub-tree\label{exmintree}}
%\vspace{-5ex}
\end{figure}

For example, the head of the minimalist tree in figure \ref{exmintree} is the leaf \textit{walks} 
(follow the direction of the projection relation in nodes and stop in a leaf).
Subtrees have their own head, for example the leaf \textit{a} is the head of the subtree $t_1$ (in figure \ref{exmintree}) and the preposition \textit{in} is the head ot $t_3$.

\subsubsection{Maximal Projection} is, for a leaf $l$, the largest subtree for which $l$ is the head.
This is the inverse notion of \textit{head}.
In the minimalist tree of figure \ref{exmintree}, the maximal projection of the leaf \textit{walks} is the full tree $t$. To describe other maximal projections in this example, the maximal projection of \textit{a} is the subtree which contains \textit{a man} and the maximal projection of the \textit{man} is the leaf \textit{man}.
In a more formal way, the maximal projection is defined as follows:

 \begin{defi}\label{defprojmax}
Let $t \in T_{MG}$, $C \in S_t$.
The \textbf{maximal projection} of $C$ (denoted by $proj_{max}(C)$) is the subtree defined by:
 \begin{itemize}
 \item $if\; C = x_1$, $proj_{max}(C) = x_1$
 \item $if \; C = C'[<(x_1,t)]$ or $C = C'[>(t, x_1)]$, $proj_{max}(C) = proj_{max}(C')$
 \item $if \; C = C'[<(t, x_1)]$ or $C = C'[>(x_1, t)]$, $proj_{max}(C) = C$
 \end{itemize}
 \end{defi}

Then $proj_{max}($\textit{walks}$) = t$. 
This logical characterization of minimalist trees and structural relations allows to prove different  properties of MG (for example that the projection is anti-symmetric), \cite{AM07th}.\\

\subsubsection{Complement and Specifier}
are relations  on subtrees with respect to the head.

Elements coming after the head provide information and they are in the \emph{complement} relation.
Let $t \in S_{MG}$, $C_1$ is a complement of $head(t) = C$, if $\projm (C) \lhd^{\ast} C_1$ and $C \prec^{+} C_1$, denoted by $C_1 \comp C$.

In the tree $t$ of figure \ref{exmintree}, the subtree $t_2$ is in a complement relation with the head \textit{walks}. It \textit{adds} information to the verb.

By contrast, elements placed before the head determine who (or what) is in the relationship.
Let $t\in S_{MG}$, $C_1$ is a specifier of $head(t) = C$, if $\projm (C) \lhd^{\ast} C_1$ and $C_1 \prec^{+} C$, denoted by $C_1 \spec C$.

In the tree $t$ of figure \ref{exmintree}, the subtree $t_1$ is in a specifier relation with the head \textit{walks}. It \textit{specifies} interpretation of the verb.

 \subsection{Minimalist Grammars\label{defgrammin}}
The computational system of MG is entirely based on features which represent linguistic properties of constituents. Rules are trigged by these features and they build minimalist trees.
A \emph{Minimalist Grammar} is defined by a quintuplet $\langle V, Features, Lex, \Phi, c \rangle$ where:
 \begin{itemize}
\item $V$ is a finite set of non-syntactic features, which contains two sets: $P$ (phonological forms, marked with / /), and $I$ (logical forms, marked with ()).

\item \textit{Features}$ = \{ B \cup S \cup L_{a} \cup L_{e}\}$ is a finite set of syntactic features,
\item $Lex$ is a set of complex expressions from $P$ and \textit{Features} (lexical items),
\item $\Phi= \{ merge , move\}$ is the set of generative rules,
\item $c \in $\textit{Features} is the feature which allows to accept derivations.
\end{itemize}

The final tree of a derivation which ends with acceptance is called a \emph{derivational tree}, which corresponds to a classical generative analysis.
Phonological forms are used as lexical items (and they could be seen as the grammar's terminal symbols).
A left-to-right reading of phonological forms in derived and accepted structures provides the recognized string.
But intermediate trees in a derivation do not stand for this. Only the derivational tree allows to recognize a string. This results from the \textit{move} rule which modifies the tree structure.
For a MG $G$ , the language $L_{G}$ recognized by $G$ is the closure of the lexicon by the generation rules.
	
 \subsection{Features\label{deftraits}}
A MG is defined by its lexicon which stores its resources.
Lexical items consist of a phonological form and a list of syntactic features. 
The syntactic set of features is divided in two subsets: one for basic categories, denoted $B$, and one for \emph{move} features, denoted $D$.
Different types of features are:

\begin{itemize}
\item $B = \{ v,\; dp, \; c, \cdots\}$ the set of \textbf{basic features}.
Elements of $B$ denote standard linguistic categories.
Note that this set contains $c$, the \emph{accepting feature} (I assume it is unique at least). 
\item $S= \{ {=}d \,\,| \,\, d \in B\}$ the set of \textbf{selectors} which expresses the necessity of another feature of $B$ of the same type
(for $d \in B$, =$d$ is the dual selector).
\item $L_a = \{ {+}k \,\,| \,\, k \in D\}$ the set of \textbf{licensors}.
These features assign an expression's property to complement another in a specifier-head relation.
\item $L_{e} = \{ {-}k \,\,| \,\,k \in D\}$ the set of \textbf{licensees}.
These features state that the expression needs to be complemented by a similar licensor.
\end{itemize}
	
Lexical sequences of features follow the syntax: $/FP/: (S(S \cup L_{a})^{\ast})^{\ast} B (L_{e})^{\ast}$

\begin{figure}[htbp]
\begin{center}
\includegraphics[height = 2 cm]{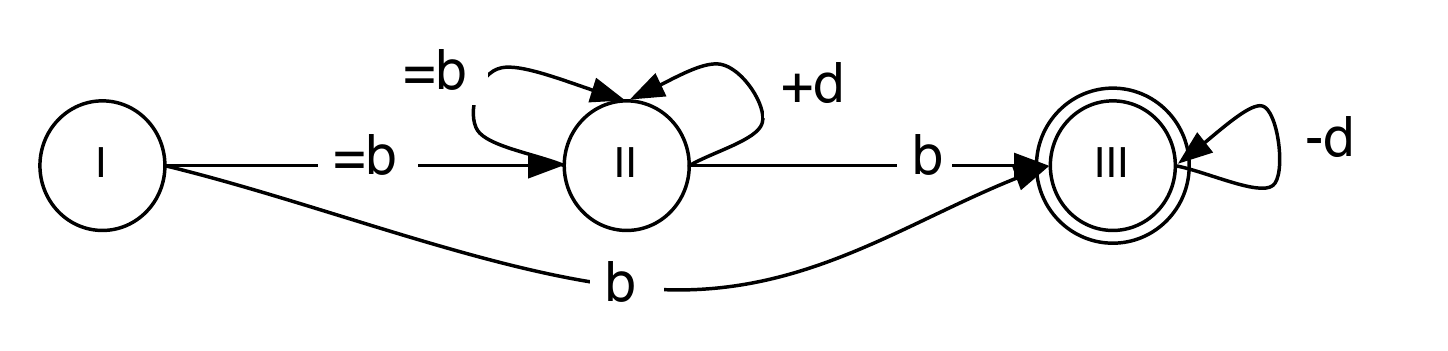}
\end{center}
\caption{Automata of acceptable sequences of features where $b \in B$ and $d \in D$\label{structureitemlexical}.}
\end{figure}

Vermaat, \cite{VW99}, proposes an automata which recognises the acceptable sequences, proposed in figure \ref{structureitemlexical}.
This structure could be divided in two parts:
the first containing a sequence of selectors and licensors (features which trigger rules, as we shall see), and the second which contains only one basic feature (the grammatical category associated to the expression) and a sequence of licensees.
The first part corresponds to stat I and II and the second to stat III and transitions to this state.
In the following, $e$ will denote any feature and $E$ a sequence of features (possibly empty).

For example, the sequence associated with an intransitive verb will be: ${=}d \;{+}case\; v$ which means that this verb must be jointed with a \textit{determinal phrase} (\textit{determinal} comes from the Generative Theory), a complex expression with feature $d$. Then it must be combined with a ${-}case$, we will see how in the next section, an then there is a structure associated with \textit{verb} (feature $v$).

Transitive verbs will extend the intransitive ones wth the list:  
$${=}d \;{+}case \; {=}d \; {+}case \; v$$
The two ${=}d$ correspond to the subject and the object of the verb. 
The first \textit{case} will be \textit{accusative} and the second \textit{nominative}. 

Another example is determiners: they are combined with a noun to build a determiner phrase and need to be unified in the structure (see the next section). Here is an example of lexicon which contains a verb, a noun and a determiner:

\begin{center}
\begin{tabular}{rll}
\textit{walks} &:& ${=}d \; {+}case\; v$\\
\textit{a} &:& ${=}n \; d \;{-case}$\\
\textit{man} &:& $n $\\ 
\end{tabular}
\end{center}

 \subsection{MG Rules\label{defregles}}
$\Phi$, the set of generating rules, contains only: \emph{merge} and \emph{move}. 
A derivation is a succession of rule applications which build trees.
These trees are partial results: the structural order of phonological forms does not need to correspond to the final one.
In the MP, a specific point, called \emph{Spell-Out} is the border between the calculus of derivations and the final result.
Rules are trigged by the feature occurring as the first element of list of features of the head.

 \subsubsection{Merge}
is the process which connects different parts.
It is an operation which joins two trees to build a new one:

 $merge$ : $T_{MG} \times T_{MG} \rightarrow T_{MG}$
 
It is triggered by a selector (${=}x$) at the top of the list of features of the head and it is realised with a corresponding basic feature ($x$) at the top of the list of features of the head of a second tree.
\emph{Merge} adds a new root which dominates both trees and cancels the two features.
The specifier/complement relation is implied by the lexical status of the tree which carried the selector. The new root node points to this tree.

\begin{figure}[htbp]
 \begin{center}
 \includegraphics[width = 6 cm, ]{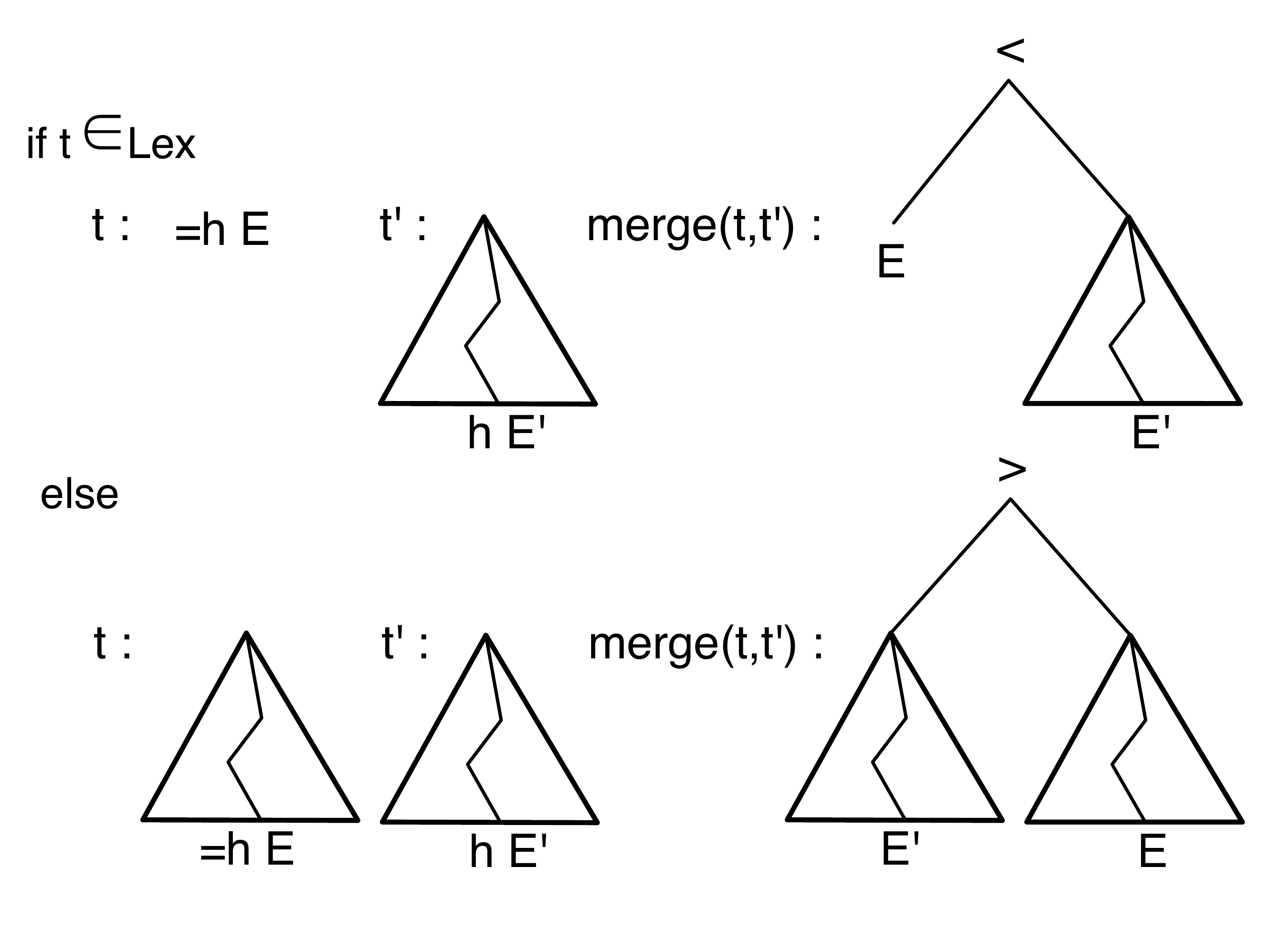}
 \includegraphics[width = 6 cm]{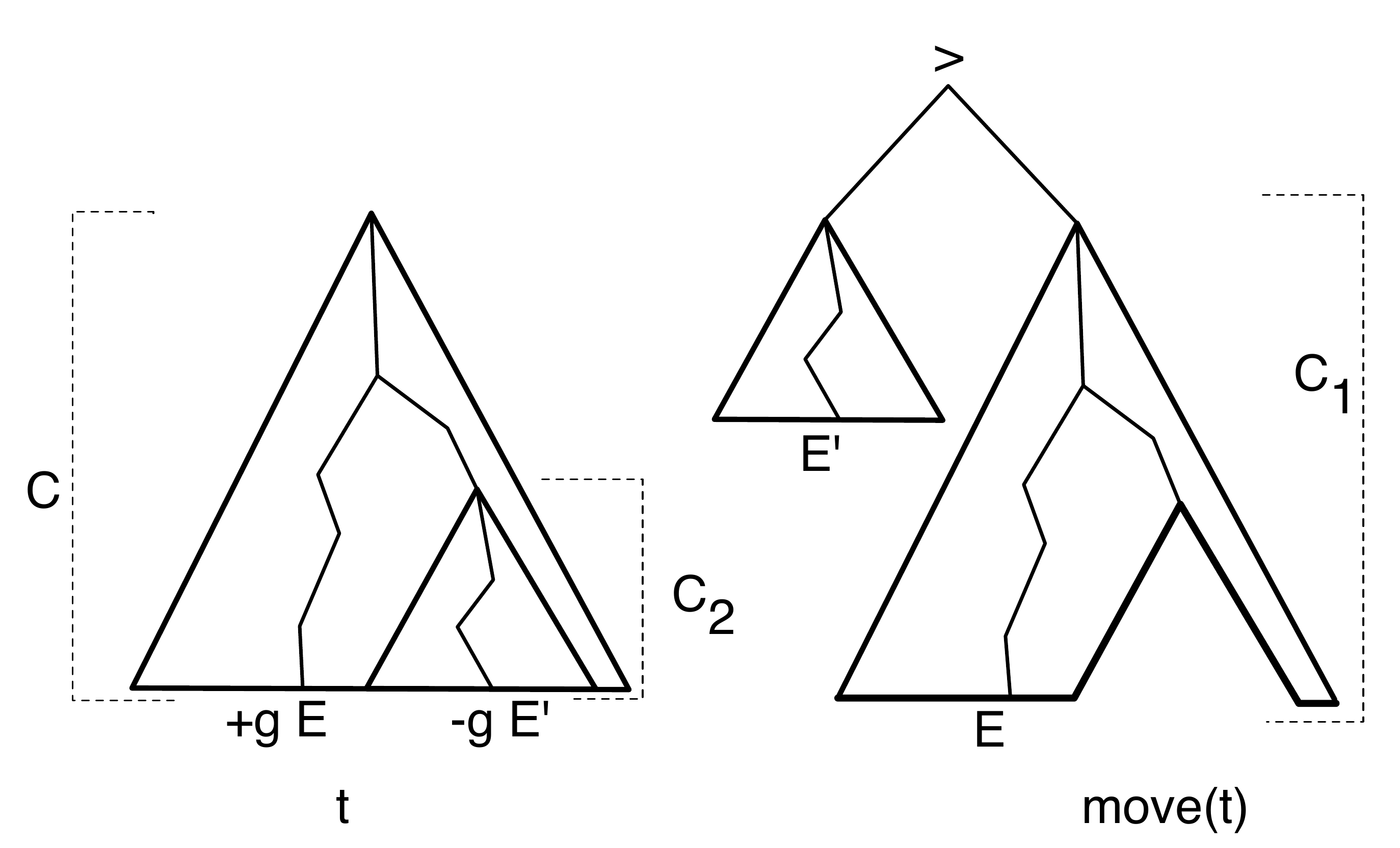}
 \caption{Tree representation of \emph{merge} and \emph{move}.\label{arbredefusion}}
 \end{center}
 \vspace{-5ex}
 \end{figure}

Let $t$,$t'$ $\in T_{MG}$ be such that $t = H_t[l : {=}h  \; E]$ and $t' = H_{t'}[l': h  \; E'] $ with $h \in B$:

$$
merge(t, t') = 
\left \{
\begin{tabular}{ll}
$<(l: \; E, H_{t'}[l': \; E'])$  & $\mathrm{if} \,
\; t \in Lex,$\\
$>(H_{t'}[l': \; E'], H_{t}[l: \; E])$ & $\mathrm{otherwise}.$\\
\end{tabular}\right.
$$
Figure \ref{arbredefusion} presents the graphical representation of \emph{merge}.

For example, to derive \textit{a man walks}, we first need to combine \textit{a} with \textit{man}, and then to combine the result with the verb:

\begin{center}
\leaf{$\;$a$\;$\\ $\cancel{{=}n} \; d \;{-}case$}
\leaf{$\;$man$\;$\\$\cancel{n}$}
\branch{2}{$<$}
\tree
and
\leaf{$\;$walks$\;$\\ $\cancel{{=}d} \; {+}case \;v$}
\leaf{$\;$a$\;$\\ $\cancel{{=}n} \; \cancel{d} \;{-}case$}
\leaf{$\;$man$\;$\\$\cancel{n}$}
\branch{2}{$<$}
\branch{2}{$<$}
\tree
\end{center}

Obtained trees do not verify the word order (only the final tree will check the right word order). In this example, the selectors are carried by lexical items, then projection relations point to the left in both cases.

\subsubsection{Move} encodes the main idea of the Minimalist Program. 
It corresponds to the movement of a constituent at the top position in a derivation.
\emph{Move} is trigged by a licensor  (${+}x$) at the top of the list of features of the head of a tree.
Then, it looks for a corresponding licensee  (${-}x$) at the top of the list of features of the head inside the tree.
If these conditions are met, the maximal projection of the node which carries the licensee is moved to the left of a new root. This node points to the right (the subtree which carries the former head). Both licensor and licensee are cancelled.
The root of the moved maximal projection is substituted by an empty leaf ($\epsilon$). This new leaf is called the \emph{trace} of the move.

Figure \ref{arbredefusion} shows a graphical representation of the move rule where the head of $C$ carries a ${+}g$ in its top features list. Then we look for a leaf with ${-}g$ in its top features list and then find its maximal projection ($C_2$) which contains all the elements which depend on it. Finally this sub-tree is moved to the left position of a new root node. Intuitively, a linguistic property is checked and the consequence is a move in first position in the tree. And strictly:

 \emph{move} : $T_{MG} \rightarrow T_{MG}$

For all tree $t = C[l: {+}g \; E, l':{-}g  \; E']$, such that $t= H_t[l:{+}g\; E]$,
there exists $C_1, C_2 \in S_t$ such that: $C_2$ is the maximal projection of the leaf $l'$ and $C_1$ is $t$ deprived of $C_2$. Then, $t = C_1[l : {+}g \; E, C_2[l' : {-}g \; E']]$ where:
 \begin{itemize}
 \item $C_2[l': {-}g \;E'] = \projm (C[l' : {-}g \;E])$
 \item $C_1[l: {+}g \; E, x_1] = \projm (C[l : {+}g \; E, x_1])$
 \end{itemize}
 $$move(t) = {>}(C_2[l':E'] , C_1[l: E, \epsilon])$$

Figure \ref{arbredefusion} presents the graphical representation of \emph{move}.

Stabler introduces some refinements to these grammars. Let us mention them. He introduces a second \emph{move}: \emph{weak move}, which does not move the phonological forms.
The precedent \emph{move} is then called \emph{strong move}, which is trigged with  capital features.
The \emph{weak move} is, like \emph{strong move}:
\begin{center}$move(t) = {>}(C_2[\epsilon :E'] , C_1[l: E, l'])$\end{center}

Variations on \emph{strong/weak} values achieve variations on phonological order.
This is an instance of the use of \emph{parameters} of the Minimalist Program.

Moreover, restrictions can be introduced on MG derivations.
An important one is the Shortest Move Condition (SMC) which blocks \emph{move} in case of ambiguity on licensees. Then, the move operation of MG with SMC is deterministic.

A locality condition could also be introduced: Specifier Island Condition - SPIC.
``Islands'' define areas which prohibit extractions.
With SPIC, a subtree cannot be moved if it is in a specifier relation within a subtree.
This condition was introduced by Stabler, in \cite{St99} drawing on works of \cite{KS00} and  \cite{Kr98}, who proposes that moved elements had to be in a complement relation.

In the previous example, the head of the last tree is the leaf \textit{walks} which contains a ${+}case$ feature as first element of its list. Then, a \textit{move} is trigged in the tree with the leaf \textit{a} which carries a  (${-}case$). The resulting tree is the following:

\begin{center}
\leaf{$\;$a$\;$\\ $\cancel{{=}n} \; \cancel{d} \;\cancel{{-}case}$}
\leaf{$\;$man$\;$\\$\cancel{n}$}
\branch{2}{$<$}

\leaf{$\;$walks$\;$\\ $\cancel{{=}d} \; \cancel{{+}case} \;v$}
\leaf{$\epsilon$}
\branch{2}{$<$}
\branch{2}{$>$}
\tree
\end{center}

The move operation modifies the position of the maximal projection of the leaf which carries the ${-}case$. The old position is substituted by an empty leaf ($\epsilon$). Finally, the tree contains only one feature which is $v$. In this small example, I did not discuss the validity of the final feature, but in a real derivation, we assume that it is not the verb which carries the ${+}case$ licensor which corresponds to the nominal case, but it is a specific item. This item corresponds to the morphological mark of the verb. Then each acceptable derivation assumes that a verb has received its time (and other properties). But exhibiting the use of this item needs other refinements of the two rules (Head-movement and Affix-Hopping).

This section did not propose a new framework for computational linguistics. This is a new definition of Stabler proposal. This way,  assumed properties of minimalist trees have been fully proved, \cite{AM07th}. Moreover this algebraic definition of MG is a perfect description to compare generated languages with other frameworks. Finally, this modifies the point of view on derivations and shows all steps of the calculus as substitution. One missing point is still the introduction of a semantic calculus. Let us now develop MCG which are defined with a syntax-semantics interface.

%\newpage
\section{Minimalist Categorial Grammars - MCG\label{gmcdef}}
In this section, we define a new Type-Theoretic Framework which is provided by the mixed calculus, a formulation of Partially Commutative Linear Logic.
It proposes to simulate MG and then keep linguistic properties of the Minimalist Program.
MCG are motivated by the syntax-semantics interface, \cite{AM07th}.
This interface, as for Lambek calculus, is based on an extension of the Curry-Howard isomorphism,
\cite{How69}.
Even though this interface is not the aim of this paper, let us discuss some important points.

The idea of encoding MP with Lambek calculus arises from \cite{LecR99} and extended versions of this work.
In these propositions, the calculus is always non-commutative, a property needed to model the left-right relation in sentences.
But the \textit{move} operation could not be defined in a proper way with non-commutative relation.
In particular, in complex utterances, the non-commutativity implies that a constituent (for example the object DP) must be fully treated before another one is introduced (for example the subject DP).
Otherwise, features are mixed and non-commutativity blocks resolutions.
It is not acceptable to normalize the framework with such a strong property and it makes the system inconsistent in regard to linguistics. 

The solution we propose is to define a new framework which allows to deal with commutative and non-commutative connectors: the mixed calculus. The main consequence on the model of this calculus is that variables in logical formulae are introduced at different places and must be unified later. In \cite{AM07th} we show how the unification is used to capture semantic phenomena which are not easily included. In few words, the idea is to consider proofs of mixed calculus as \textit{phases} of a verb. Phases have been introduced by Chomsky to detail different modifications which occur on a verb. Several linguists have showed that phases have implications on semantics, for example the theta-roles must be allocated after a specific phase. This is exactly the result of the syntax-semantics interface of MCG. Full explanations need more space to be presented, but the main contribution of MCG is to propose an efficient syntax-semantics interface in the same perspective as MG.

In this section, we will detail MCG and expose their structural link with MG. First we present the mixed calculus, then we give definitions of MCG and show proofs of the mixed calculus produced by MCG (together with their linguistic properties).

\subsection{Mixed calculus}
MCG are provided with mixed calculus, \cite{Gro96}, a formulation of Partially Commutative Linear Logic. 
Hypotheses are either in a non-commutative order ($<;>$) or in a commutative one ($(,)$)
The plain calculus contains introduction and elimination rules for:
\begin{itemize}
\item the non-commutative product $\odot$:

\smallskip

\hfill
\prooftree
	\Delta \vdash  A \odot B
	\quad \Gamma, <A; B>, \Gamma ' \vdash C
	\justifies
	\Gamma, \Delta , \Gamma '  \vdash C 
	\using
             [\odot_{e}]
         \thickness=0.07em
\endprooftree
\hfill 
\prooftree
	\Delta \vdash A
	\quad \Gamma \vdash B
	\justifies
	<\Delta; \Gamma> \vdash A \odot B
	\using
             [\odot_{i}]
         \thickness=0.07em
\endprooftree
\hfill

\item its residuals ($\lfrom$ and $\lto$):

\smallskip

\hfill
\prooftree
	\Gamma \vdash A
	\quad \Delta \vdash  A \backslash C
	\justifies
	<\Gamma; \Delta> \vdash C 
	\using
             [\backslash_{e}]
         \thickness=0.07em
\endprooftree
\hfill
\prooftree
	\Delta \vdash  A / C
	\quad \Gamma \vdash A
	\justifies
	< \Delta ; \Gamma> \vdash C 
	\using
             [/_{e}]
         \thickness=0.07em
\endprooftree
\hfill

\bigskip 

\hfill
\prooftree
	<A; \Gamma> \vdash C
	\justifies
	\Gamma \vdash A \backslash C 
	\using
             [\backslash_{i}]
         \thickness=0.07em
\endprooftree
\hfill
\prooftree
	<\Gamma; A> \vdash  C
	\justifies
	\Gamma \vdash C/A 
	\using
             [/_{i}]
         \thickness=0.07em
\endprooftree
\hfill

\item the commutative product $\otimes$:

\smallskip

\hfill
\prooftree
	\Delta \vdash  A \otimes B
	\quad \Gamma, (A, B),  \Gamma ' \vdash C
	\justifies
	\Gamma, \Delta , \Gamma '  \vdash C 
	\using
             [\otimes_{e}]
         \thickness=0.07em
\endprooftree
\hfill
\prooftree
	\Delta \vdash A
	\quad \Gamma \vdash B
	\justifies
	(\Delta, \Gamma) \vdash A \otimes B
	\using
             [\otimes_{i}]
         \thickness=0.07em
\endprooftree
\hfill

\item its residual $\multimap$:

\smallskip

\hfill
\prooftree
	 \Gamma \vdash A
	\quad \Delta \vdash  A \multimap C
	\justifies
	(\Gamma, \Delta) \vdash C 
	\using
             [\multimap_{e}]
         \thickness=0.07em
\endprooftree
\hfill
\prooftree
	(A, \Gamma ) \vdash  C
	\justifies
	\Gamma \vdash A \multimap C 
	\using
             [\multimap_{i}]
         \thickness=0.07em
\endprooftree
\hfill

\end{itemize}

The product connectors of the mixed calculus use in a first step hypotheses to mark positions in the proof and in a second one substitute the result of an another proof in these positions using a product elimination (the commutative/non-commutative status depends on relations between hypotheses).
This is exactly the process we will use to define the \textit{move} rule of MCGs.

Moreover, the calculus contains an axiom rule and an entropy rule. This last one allows to relax the order between hypotheses. We will use this rule to define \textit{merge} in MCG as we will see in the following section.

\smallskip

\hfill 
\prooftree
	\justifies
	A \vdash A
	\using
             [axiom]
         \thickness=0.07em
\endprooftree
\hfill 
\prooftree
	\Gamma \vdash C
	\justifies
	\Gamma' \vdash C 
	\using
             [\mbox{entropy --- whenever }\Gamma'\sqsubset\Gamma]
         \thickness=0.07em
\endprooftree
\hfill 

\smallskip

This calculus has been shown to be normalizable, \cite{MR07} and derivations of MCG will be proofs of the mixed calculus in normal form.

\subsection{Minimalist Categorial Grammars\label{defgmc}}
As MG, MCG are lexicalized grammars. Derivations are led by formulae associated with lexical items built with connectors of the mixed logic.
They are specific proofs of the mixed logic, labelled to realise the phonological and semantic tiers.
Phonological labels on proofs will be presented with definitions of MCG rules.

A MCG is defined by a quintuplet
$\langle N, $\textsc{p}$, Lex, \Phi, C \rangle$ where :
\begin{itemize}
\item $N$ is the union of two finite disjoint sets $Ph$ and $I$ which are respectively the \textit{set of phonological forms} and the one of \textit{logical forms}.

\item \textsc{p} is the union of two finite disjoint sets \textsc{p}$_1$ and \textsc{p}$_2$ which are respectively the \textit{set of constituent features} (the set B of MG) and the one of \textit{move features} (the set D of MG).

\item $Lex$ is a finite subset of $E \times F \times I$, the set of lexical items \footnote{In the following, $Lex$ is a subset of $E \times F$. The semantic part is used for the syntax-semantics interface which is not detailed here.}.

\item $\Phi= \{ merge , move\}$ is the set of generative rules,
\item $C \in$ \textsc{p} is the accepting formulae.
\end{itemize}

As mentioned in the previous section, \textit{move} is defined using a product elimination. In MG, a constituent is first introduced in a tree using its basic feature and then can be moved using its licensees. In MCG, a constituent will be introduced only when all its positions (which correspond to the basic feature and its licensees) have been marked in the proof by specific hypotheses. But we need to distinguish the type of the basic feature from the licensees features.
That is why \textsc{p} is divided in two subsets \textsc{p$_1$} and \textsc{p$_2$}. This sub-typing of formulae is used to well define lexicons of MCG.

The set $E$ is $Ph^{\ast}$, and the set $F$, the set of formulae used to build $Lex$, is defined with the set $\textsc{p}$, the commutative product $\otimes$ and the two non-commutative implications $\lfrom$ and $\lto$.
Formulae of $F$ are recognized by the non-terminal \textsc{l} of the following grammar:

$$
\begin{array}{l}
\textsc{l} ::= (\textsc{b}) \lfrom \textsc{p}_1\; |\; \textsc{c}\\
\textsc{b} ::= \textsc{p}_1 \lto (\textsc{b})\; | \textsc{p}_2 \lto (\textsc{b})\; |\;\; \textsc{c}\\
\textsc{c} ::= \textsc{p}_2 \otimes (\textsc{c})\; |\; \textsc{c}_1\\
\textsc{c}_1 ::= \textsc{p}_1\\
\end{array}
$$

In more details, MCG formulae start with a $\lfrom$ which is followed by a sequence of $\lto$.
This sequence contains operators allowing to compose the proof with another one (operators are the translation of selectors and licensors).
Lexical formulae are ended by a sequence of $\otimes$.
To sum up, these formulae have the structure $(c_m \backslash \ldots \backslash c_1 \backslash (b_1 \otimes \ldots \otimes b_n \otimes a ) ) / d$, with $a \in$ \textsc{p}$_1$, $b_i \in$ \textsc{p}$_2$, $c_j \in$  \textsc{p} and $d \in$  \textsc{p}$_1$.
This structure corresponds to the two parts of the list of features we have mentioned in the previous section.

For the example \textit{a man walks}, the MCG's lexicon is the following:

$$\begin{array}{rll}
$walks$&:&case \backslash v/d\\
$a$&:& (case \otimes d) / n\\
$man$&:&n\\
\end{array}$$

Licensees, which express the need for an information, are there seen as a specific part of the basic feature (a part of the main sub-type). Licensors will be cancelled with an hypothesis to mark a position in the proof.
Distinction between them is not written by an \textit{ad hoc} marker but by structural relations inside the formula.
Before we explain the \textit{move} and \textit{merge} rules, let us present the phonological tiers.
\subsection{Derivations}

\subsubsection{Labels.}
Derivations of MCG are labelled proofs of the mixed calculus.
Before defining labelling, we define labels and operations on them.

Let $V$ be an uncountable and finite set of variables such that: $Ph \cap V = \emptyset $.
$T$ is the union of $Ph$ and $V$. We define the set $\Sigma$, called \emph{labels set} as the set of triplets of elements of $T^\ast$. Every position in a triplet has a linguistic interpretation: they correspond to specifier/head/complement relations of minimalist trees.
A label $r$ will be considered as $r=(r_{spec}, r_{head}, r_{comp})$.

For a label in which there is an empty position, we adopt the following notation: $r_{-head} = (r_{spec}, \epsilon, r_{comp})$, $r_{-spec} = (\epsilon, r_{head}, r_{comp})$, $r_{-comp} = (r_{spec}, r_{head}, \epsilon)$.
We introduce variables in the string triplets and a substitution operation. They are used to modify a position inside a triplet by a specific material. Intuitively, this is the counterpart in the phonological calculus of the product elimination.
The set of variables with at least one in $r$ is denoted by $Var(r)$.
The number of occurrences of a variable $x$ in a string $s \in T^\ast$ is denoted by $|s|_{x}$, and the number of occurrences of $x$ in $r$ by $\varphi_x (r)$.
A label is \emph{linear} if for all $x$ in $V$, $\varphi_x(r) \leqslant 1$.

A \emph{substitution} is a partial function from $V$ to $T^\ast$.
For $\sigma$ a substitution, $s$ a string of $T^{\ast}$ and $r$ a label, we note $s.\sigma$ and $r.\sigma$ the string and the label obtained by the simultaneous substitution in $s$ and $r$ of the variables by the values associated by  $\sigma$ (variables for which $\sigma$ is not defined remain the same).

If the domain of definition of a substitution $\sigma$ is finite and equal to $x_1, \dots, x_n$ and $\sigma(x_i) = t_i$, then $\sigma$ is denoted by $[t_1/x_1, \dots , t_n/x_n]$. Moreover, for a sequence $s$ and a label $r$, $s. \sigma$ and $r. \sigma$ are respectively denoted $s[t_1/x_1, \dots , t_n/x_n]$ and $r[t_1/x_1, \dots , t_n/x_n]$.
Every injective substitution which takes values in $V$ is called \emph{renaming}.
Two labels $r_1$ and $r_2$ (respectively two strings $s_1$ and $s_2$) are equal modulo a renaming of variables if there exists a renaming $\sigma$ such that $r_1.\sigma = r_2$ (\resp $s_1.\sigma = s_2$).

Finally, we need another operation on string triplets which allows to combine them together: the string concatenation of $T^\ast$ is noted $\bullet$.
Let $Concat$ be the operation of concatenation on labels which concatenates the three components in the linear order: for  $r \in \Sigma$, $Concat (r) = r_{spec} \bullet r_{head} \bullet r_{comp}$.

We then have defined a phonological structure which encodes specifier/comple\-ment/head relations and two operations (substitution and concatenation). These two operations will be counterparts in the phonological calculus of \textit{merge} and \textit{move}.

\subsubsection{Labelled proofs.}
Before exhibiting the rules of MCG, the concept of labelling on a subset of rules of the mixed logic is introduced.
\emph{Minimalist logic} is the fragment of mixed logic composed by the axiom rule, $\lto_e$, $\lfrom_e$, $\otimes_e$ and $\sqsubset$.

For a given MCG $G = \langle N, $\textsc{p}$, Lex, \Phi, C \rangle$, let a \emph{$G$-background} be $x: A$ with $x\in V$ and $A \in F$, or $\langle G_1 ; G_2\rangle$ or else $(G_1 , G_2)$ with $G_1$ and $G_2$ some \emph{$G$-backgrounds} which are defined on two disjoint sets of variables.
$G$-backgrounds are series-parallel orders on subsets of $V \times F$.
They are naturally extended to the entropy rule, noted $\sqsubset$.
A \emph{$G$-sequent} is a sequent of the form:
$\Gamma \seq_G (r_s, r_t, r_c) : B$ 
where $\Gamma$ is a $G$-background,
$B \in F$ and $( r_s, r_t, r_c ) \in \Sigma$.

A \emph{$G$-labelling} is a derivation of a $G$-sequent obtained with the following rules:

\begin{center}
\prooftree
	\langle s, A \rangle \in Lex
	\justifies
	\seq_G (\epsilon, s, \epsilon) : A
	\using [Lex]
\endprooftree

\bigskip

\prooftree
	x \in V
	\justifies
	x: A \seq_G (\epsilon, x, \epsilon) : A
	\using [axiom]
\endprooftree

\bigskip

\prooftree
	\Gamma \seq_G r_1 : A \lfrom B
	\quad \Delta \seq_G r_2 : B
	\quad Var(r_1) \cap Var(r_2) = \emptyset
	\justifies
	\langle \Gamma ; \Delta\rangle \seq_G (r_{1s},r_{1t},r_{1c} \bullet Concat(r_2)) : A
	\using [\lfrom_e]
\endprooftree

\bigskip

\prooftree
	\Delta \seq_G r_2 : B
	\quad \Gamma \seq_G r_1 : B \lto A
	\quad Var(r_1) \cap Var(r_2) = \emptyset
	\justifies
	\langle \Gamma ; \Delta\rangle \seq_G  (Concat(r_2)\bullet r_{1s},r_{1t},r_{1c}) : A
	\using [\lto_e]
\endprooftree

\bigskip

\prooftree
	\Gamma \seq_G r_1 : A \otimes B
	\quad \Delta[x: A , y:B] \seq_G r_2 : C
	\quad Var(r_1) \cap Var(r_2) = \emptyset
	\quad A \in \mbox{\textsc{p}}_2
	\justifies
	\Delta[ \Gamma] \seq_G  r_2[Concat(r_1)/x , \epsilon/y ] : C
	\using [\otimes_e]
\endprooftree

\bigskip

\prooftree
	\Gamma \seq_G r : A
	\quad \Gamma' \sqsubset \Gamma
	\justifies
	\Gamma' \seq_G r : A
	\using [\sqsubset] 
\endprooftree
\end{center}

Note that a $G$-labelling is a proof tree of the minimalist logic on which sequent hypotheses are decorated with variables and sequent conclusions  are decorated with labels. Product elimination is used with a substitution on labels and implication connectors with concatenation (a triplet is introduced in another one by concatenating its three components).

If $\Gamma \seq_G r : B$ is a $G$-sequent derivable, then $r$ is linear, and $Var(r)$ is exactly the set of variables in $\Gamma$.
Finally, for all renamings $\sigma$, $\Gamma.\sigma \seq_G r.\sigma : B$ is a $G$-sequent differentiable.\\

\subsubsection{\emph{Merge} and \emph{Move} rules}
are simulated by combinations of rules of the minimalist logic producing $ G $-labeling.\\

\textbf{\textit{Merge}} is the elimination of $\lfrom$ (\resp $\lto$) immediately followed by an \emph{entropy} rule.
The meaning of this rule is joining two elements in regard to the left-right order (then non-commutative connectors are used) and, as mentioned earlier, all hypotheses must be accessible. To respect this, a commutative order between hypotheses is needed. Then an entropy rule immediately follows each implication elimination.

For the phonological tier, a label is concatenated in the complement (respectively specifier) position in another one.
Note that a \emph{merge} which uses $\lfrom$ must be realized with a lexical item, so the context is always empty.
$$
\displaylines{
\prooftree
	\prooftree
		\seq ( r_{spec}, r_{head}, r_{comp}) : A\lfrom B
		\quad \Delta \seq s : B
		\justifies
		 \Delta \seq ( r_{spec}, r_{head}, r_{comp}\bullet Concat(s)) : A
		\using
		  [\lfrom_e]
	\endprooftree
	\justifies
	\Delta \seq ( r_{spec}, r_{head}, r_{comp}\bullet Concat(s)) : A
	\using
	  [\sqsubset]
\endprooftree
}$$

$$
\displaylines{
\prooftree
	\prooftree
		 \Delta \seq s : B
		\quad \Gamma \seq (r_{spec}, r_{head}, r_{comp}) : B \lto A
		\justifies
		\langle \Delta ; \Gamma \rangle \seq (Concat(s)\bullet r_{spec}, r_{head}, r_{comp}) : A
		\using
		  [\lto_e]
	\endprooftree
	\justifies
	\Delta , \Gamma \seq (Concat(s)\bullet r_{spec}, r_{head}, r_{comp}) : A
	\using
	  [\sqsubset]
\endprooftree
}
$$
These combinations of rules are noted $[mg]$.

For example, the proof of the utterance \textit{a man walks} begins with the formulae of \textit{walks}:
$case \backslash v/d$. The first step of the calculus is to introduce two hypotheses, one for $d$ and the other for $case$. The result is the following proof:

$$
\displaylines{
\prooftree
	\quad v : case \seq (\epsilon, v,\epsilon) : case
	\prooftree
		\seq (\epsilon, walks, \epsilon) : case \backslash v/d
		\quad u : d \seq (\epsilon, u,\epsilon) : d
		\justifies
		u:d \seq (\epsilon, walks, u) : case \backslash v
		\using
		  [mg]
	\endprooftree
	\justifies
	(v: case, u:d) \seq (\epsilon, walks, u) : v
	\using
	  [mg]
\endprooftree
}
$$

In parallel, the derivation joins the determiner \textit{a} and the noun \textit{man}:

$$
\displaylines{
\prooftree
	\seq (\epsilon, a, \epsilon) : (case \otimes d) / n
	\quad \seq (\epsilon, man, \epsilon) : n
	\justifies
	\seq (\epsilon, a, man) : case \otimes d
	\using
	  [mg]
\endprooftree
}
$$

Note that the first proof contains two hypotheses which correspond to the type of the main formula in the second proof.
The link between these two proofs will be made by a \textit{move}, as we will show later.\\

\textbf{\textit{ Move}} is simulated by an elimination of a commutative product in a proof and, for the phonological calculus, is a substitution. We have structured the lexicons and the merge rule to delay to the move rule only the substitution part of the calculus.

$$
\displaylines{
\prooftree
	\quad \Gamma \seq r_1 : A \otimes B
	\quad  \Delta[u : A, v: B] \seq r_2 : C
	\justifies
	 \Delta[\Gamma] \seq r_2[Concat(r_1) / u , \epsilon /  v] : C
	\using
	  [\otimes_e]
\endprooftree}
$$
\smallskip

This rule is applied only if $A \in$ \textsc{p}$_2$ and $B$ is of the form $B_1 \times \ldots B_n \times D$ where $B_i \in $ \textsc{p}$_2$ and $D\in $  \textsc{p}$_1$.

This rule is noted $[mv]$.
\emph{Move} uses hypotheses as resources. 
The calculus places hypotheses in the proof, and when all hypotheses corresponding to a constituent are introduced, this constituent is substituted.
The hypothesis \textsc{p}$_1$ is the first place of a moved constituent and hypotheses of \textsc{p}$_2$ mark the different places where the constituent is moved or have a trace.

In recent propositions, Chomsky proposes to delay all moves after the realisation of all merges. MCG could not encode this but contrary to MG where a \textit{move} blocks all the process, in MCG \textit{merge} could happen, except  in the case of hypotheses of a given constituent shared by two proofs which must be linked by a \textit{move}. 

In our example, we have two proofs:
\begin{itemize}
\item one for the verb: $(v: case, u:d) \seq (\epsilon, walks, u) : v$
\item one for the DP: $\seq (\epsilon, a, man) : case \otimes d$
\end{itemize}

The first hypothesis corresponds to the entry position of the DP in MG and the second to the moved position. Here, we directly introduce the DP by eliminating the two hypotheses in the same step:

$$
\displaylines{
\prooftree
	 \seq (\epsilon, a, man) : case \otimes d
	\quad (v: case, u:d) \seq (\epsilon, walks, u) : v
	\justifies
	\seq (a \;man, walks, \epsilon) : v
	\using
	  [mv]
\endprooftree}
$$
\smallskip

The phonological result is \textit{a man walks}. The proof encodes the same structure as the derivational tree of MG (modulo a small transduction on the proof).

For \emph{cyclic move} (where a constituent is moved several times) all hypotheses inside this move must be linked together upon their introduction in the proof.
For this, when a new hypothesis $A$ is introduced, a $[mv]$ is applied with a sequent with hypothesis $A\otimes B \seq A\otimes B$ where $A$ is in \textsc{p$_2$} and $B$ is of the form $B_1 \otimes \ldots \otimes B_n \otimes D$ where $B_i \in $ \textsc{p}$_2$ and $D\in $  \textsc{p}$_1$.

$$
\displaylines{
\prooftree
	\quad x : A \otimes B \seq (\epsilon, x, \epsilon) : A \otimes B
	\quad \Delta[u : A, v: B] \seq r : C
	\justifies
	\Delta[A \otimes B] \seq r[x / u , \epsilon /  v] : C
	\using
	  [\otimes_e]
\endprooftree}
$$

\smallskip

In the definition of \emph{merge}, the systematic use of entropy comes from the definition of \emph{move}.
As it was presented, \emph{move} consumes hypotheses of the proof. But, from a linguistic perspective, these hypotheses could not be supposed introduced next to each other. The non-commutative order inferred from $\lto_e$ and $\lfrom_e$ blocks the \emph{move} application. To avoid this, the entropy rule places them in commutative order.
In MCG, all hypotheses are in the same relation, then to simplify the reading of proofs, the order is denoted only with '$,$'.

The strong/weak move could be simulated with the localization of the substitution (if hypotheses are in \textsc{p}$_1$ or \textsc{p}$_2$).
\smallskip

\begin{center}
\prooftree
	 s : \Gamma \seq A \otimes B
	\quad r[u,v] : \Delta[u : A, v: B] \seq C
	\justifies
	r[Concat(s) / u , \epsilon /  v] : \Delta[\Gamma] \seq C
	\using
	  [move_{strong}]
\endprooftree

\bigskip

\prooftree
	s : \Gamma \seq A \otimes B
	\quad r[u,v] : \Delta[u : A, v: B] \seq C
	\justifies
	r[\epsilon / u , Concat(s) /  v] : \Delta[\Gamma] \seq C
	\using
	  [move_{weak}]
\endprooftree
\end{center}
\smallskip

This version of move is quite different from the one presented for MG, but is close to one developed for later MG such as \cite{KB06}.

The main difference between MG and MCG comes from \emph{move}: in MCG, constituents do not move but use hypotheses marking their places. MCG uses commutativity properties of mixed logic and see hypotheses as resources.
To sum up, the derivation rules of MCG is the following set of rules:
\begin{center}
\prooftree
	\langle s, A \rangle \in Lex
	\justifies
	\seq_G (\epsilon, s, \epsilon) : A
	\using [Lex]
\endprooftree
\qquad
\prooftree
	\seq ( r_{spec}, r_{head}, r_{comp}) : A\lfrom B
	\quad \Delta \seq s : B
	\justifies
	\Delta \seq ( r_{spec}, r_{head}, r_{comp}\bullet Concat(s)) : A
	\using
	  [mg]
\endprooftree

\bigskip

\prooftree
	 \Delta \seq s : B
	\quad \Gamma \seq (r_{spec}, r_{head}, r_{comp}) : B \lto A
	\justifies
	\Delta , \Gamma \seq (Concat(s)\bullet r_{spec}, r_{head}, r_{comp}) : A
	\using
	  [mg]
\endprooftree

\bigskip

\prooftree
	\quad \Gamma \seq r_1 : A \otimes B
	\quad  \Delta[u : A, v: B] \seq r_2 : C
	\justifies
	 \Delta[\Gamma] \seq r_2[Concat(r_1) / u , \epsilon /  v] : C
	\using
	  [mv]
\endprooftree

\bigskip
\end{center}

The set $\mathbb{D}_G$ of recognized derivations by a MCG $G$ is the set of proofs obtained with this set of rules and for which the concluding sequent is $\seq r : C$.
The language generated by $G$ is $L(G) = \{ Concat(r) | \seq r : C \in \mathbb{D}_G\}$.

These derivations do not formally conserve the projection relation (nor the specifier, head and complement relations).
These principles are reintroduced with strings.
However, the head of a proof could be seen as the \emph{principal formula} of mixed logic, and then by extension, the maximal projection is the proof for which a formula is the principal one. Specifier and complement are only elements on the right or left of this formula.

An interesting remark is that rules of MCG do not use the introduction rule of the mixed calculus. This way, they only try to combine together formulae extracted from a lexicon and hypotheses.
As in MG where a derivation cancels features, the MCG system only consumes hypotheses and always reduces the size of the main  formula (only the size of the context could increase). This corresponds to the cognitive fact that we stress the system in the analysis perspective.
Introduction rules could be seen as captured by the given lexicon. But, because of the strong structure of the items, we directly associate formulae and strings.

We have presented all the MCG rules and lexicon, and illustrated them with a tiny example which encodes the main properties of this framework.

\section{Conclusion}
In this article, we propose new definitions of MG based on an algebraic description of trees.
These definitions allow to check properties of this framework and moreover give a formal account to analyse links with other frameworks.
Then, we give the definitions of MCG, a Type-Theoretic framework for MG. In this framework, \emph{merge} and \emph{move} are simulated by rules of the mixed logic (an extension of Lambek calculus to product and non-commutative connectors).
The phonological calculus is added by labelling proofs of this logic.
%The semantic tiers is given in \cite{AM07th}.

The main contribution of MCG is certainly its syntax-semantics interface. This calculus is synchronized on proofs of MCG. But more technical details are needed to present this interface and the linguistic properties which it encodes.
We delay the presentation of this interface to a future presentation.

Finally, the syntax-semantics interface of MCG should be used under the condition they keep properties of MG.
This is the aim of another future article which will present the proof of inclusion of MG generated languages in MCG generated languages. To prove this property, two alternative representations of MG and MCG derivations are introduced: \emph{alternative derived structures} and \emph{split proofs} and the corresponding \emph{merge} and \emph{move}. These structures and rules make the gap between the two kinds of derivations.
They need technical details and more space to be presented.

Definitions and proofs could be easily extended to refinements of \emph{merge}: \emph{Affix-Hopping} and \emph{Head-Movement} because these operations derived the same strings in both structures.
But we have not included these rules in this presentation.
On another hand, the proof of inclusion presented here does not include the SMC.
The interpretation of SMC in MCG must be better defined before being included in such perspective.
The generative power of these grammars with shortest move condition is still open.

This article is a first step to several perspectives which make a strong link between a well defined framework with many linguistic properties and a new one which captures this framework and proposes a syntax-semantics interface.

\section*{Acknowledgements}
The author would like to express his deep gratitude to his supervisors Alain Lecomte and Christian Retor\'e.
In particular, discussions with Alain Lecomte was a source of supports and good advices which turn the author to this research.

The author also wants to thank the associated editors and the anonymous reviewers for their constructive remarks and suggestions, 
and finally Patrick Blackburn and Mathieu Morey for their careful readings.

\bibliographystyle{splncs}
\bibliography{articlealain}

\begin{thebibliography}{10}

\bibitem{Chom95}
Chomsky, N.:
\newblock The Minimalist Program.
\newblock MIT Press, Cambridge (1995)

\bibitem{Sta97}
Stabler, E.:
\newblock Derivational minimalism.
\newblock Logical Aspect of Computational Linguistic \textbf{1328} (1997)

\bibitem{AM07th}
Amblard, M.:
\newblock Calcul de repr{\'e}sentations s{\'e}mantiques et suntaxe
  g{\'e}n{\'e}rative: les grammaires minimalistes cat{\'e}gorielles.
\newblock PhD thesis, universit{\'e} de Bordeaux 1 (septembre 2007)

\bibitem{RMusk03}
Muskens, R.:
\newblock {L}anguage, {L}ambdas, and {L}ogic.
\newblock In Kruijff, G.J., Oehrle, R., eds.: Resource Sensitivity in Binding
  and Anaphora. Studies in Linguistics and Philosophy.
\newblock Kluwer (2003)  23--54

\bibitem{PdG01b}
de~Groote, P.:
\newblock Towards abstract categorial grammars.
\newblock Association for Computational Linguistics, 39th Annual Meeting and
  10th Conference of the European Chapter, Proceedings of the Conference (2001)

\bibitem{Polletal09}
Mansfield, L., Martin, S., Pollard, C., Worth, C.:
\newblock Phenogrammatical labelling in convergent grammar: the case of wrap.
\newblock unpublished (2009)

\bibitem{BE96}
Berwick, R., Epstein, S.:
\newblock On the convergence of 'minimalist' syntax and categorial grammars
  (1996)

\bibitem{RLC}
Retor\'e, C., Stabler, E.:
\newblock Reseach on Language and Computation. Volume 2(1).
\newblock Christian Retor\'e and Edward Stabler (2004)

\bibitem{Lec04}
Lecomte, A.:
\newblock Rebuilding the minimalist program on a logical ground.
\newblock Journal of Research on Language and Computation \textbf{2(1)} (2004)
  27--55

\bibitem{tC04}
Cornell, T.:
\newblock Lambek calculus for transformational grammars.
\newblock Journal of Research on Language and Computation \textbf{2(1)} (2004)
  105--126

\bibitem{LecR99}
Lecomte, A., Retor{\'e}, C.:
\newblock Towards a logic for minimalist.
\newblock Formal Grammar (1999)

\bibitem{LR01acl}
Lecomte, A., Retor{\'e}, C.:
\newblock Extending {L}ambek grammars: a logical account of minimalist
  grammars.
\newblock In: Proceedings of the 39th Annual Meeting of the {A}ssociation for
  {C}omputational {L}inguistics, {ACL} 2001, Toulouse, {ACL} (July 2001)
  354--361

\bibitem{AL05}
Lecomte, A.:
\newblock Categorial grammar for minimalism.
\newblock Language and Grammar : Studies in Mathematical Linguistics and
  Natural Language \textbf{CSLI Lecture Notes}(168) (2005)  163--188

\bibitem{ALR03}
Amblard, M., Lecomte, A., Retor{\'e}, C.:
\newblock Syntax and semantics interacting in a minimalist theory.
\newblock Prospect and advance in the Syntax/Semantic interface (October 2003)
  17--22

\bibitem{ALR04}
Amblard, M., Lecomte, A., Retor{\'e}, C.:
\newblock Synchronization syntax semantic for a minimalism theory.
\newblock Journ{\'e}e S{\'e}mantique et Mod{\'e}lisation (mars 2004)

\bibitem{journals/jfp/Huet97}
Huet, G.P.:
\newblock The zipper.
\newblock J. Funct. Program \textbf{7}(5) (1997)  549--554

\bibitem{LC}
Levy, J.J., Cori, R.:
\newblock Algorithmes et Programmation.
\newblock Ecole Polytechnique

\bibitem{Chomsky73}
Chomsky, N.:
\newblock Conditions on transformations.
\newblock In Kiparsky, S.A..P., ed.: A Festschrift for Morris Halle.
\newblock Holt Rinehart and Winston (1973)  232--286

\bibitem{VW99}
Vermaat, W.:
\newblock Controlling movement: Minimalism in a deductive perspective.
\newblock Master's thesis, Universiteit Utrecht (1999)

\bibitem{St99}
Stabler, E.:
\newblock Remnant movement and structural complexity.
\newblock Constraints and Resources in Natural Language Syntax and Semantics
  (1999)  299--326

\bibitem{KS00}
Koopman, H., Szabolcsi, A.:
\newblock A verbal Complex.
\newblock MIT Press, Cambridge (2000)

\bibitem{Kr98}
Kayne, R.S.:
\newblock Overt vs covert movment.
\newblock Syntax 1,2 (1998)  128--191

\bibitem{How69}
Howard, W.A.:
\newblock The formulae-as-types notion of construction.
\newblock In Hindley, J., Seldin, J., eds.: To H.B. Curry: Essays on
  Combinatory Logic, $\lambda$-calculus and Formalism.
\newblock Academic Press (1980)  479--490

\bibitem{Gro96}
de~Groote, P.:
\newblock Partially commutative linear logic: sequent calculus and phase
  semantics.
\newblock In Abrusci, V.M., Casadio, C., eds.: Third Roma Workshop: Proofs and
  Linguistics Categories -- Applications of Logic to the analysis and
  implementation of Natural Language, Bologna:CLUEB (1996)  199--208

\bibitem{MR07}
Amblard, M., Retore, C.:
\newblock Natural deduction and normalisation for partially commutative linear
  logic and lambek calculus with product.
\newblock Computation and Logic in the Real World, CiE 2007 \textbf{Quaderni
  del Dipartimento di Scienze Matematiche e Informatiche "Roberto Magari"}
  (june 2007)

\bibitem{KB06}
Kobele, G.:
\newblock Generating Copies: An Investigation into Structural Identity in
  Language and Grammar.
\newblock PhD thesis, University of California, Los Angeles (2006)

\end{thebibliography}
\end{document}